%% file: KurdishPOST.tex
\documentclass[11pt]{article}
\usepackage{arabtex}
\usepackage{coling2018}
\usepackage{utf8}
\setcode{utf8}
\usepackage{url}
\usepackage{graphicx} 
\usepackage{tikz}
\usepackage{caption} 
\usepackage[stable]{footmisc}
\date{}

\begin{document}

\title{\textbf{Part of Speech Tagging (POST) of a Low-resource Language using another Language \\ {\small{\textit{Developing a POS-Tagged Lexicon for Kurdish (Sorani) using a Tagged Persian (Farsi) Corpus}}}}}

\author{
	\begin{tabular}[t]{c}
		Hossein Hassani\\
		\textnormal{University of Kurdistan Hewl\^er}\\
		\textnormal{Kurdistan Region - Iraq}\\
		{\tt hosseinh@ukh.edu.krd}
	\end{tabular}
}

\maketitle

\begin{abstract}

Tagged corpora play a crucial role in a wide range of Natural Language Processing. The Part of Speech Tagging (POST) is essential in developing tagged corpora. It is time-and-effort-consuming and costly, and therefore, it could be more affordable if it is automated. The Kurdish language currently lacks publicly available tagged corpora of proper sizes. Tagging the publicly available Kurdish corpora can leverage the capability of those resources to a higher level than what raw or segmented corpora can provide. Developing POS-tagged lexicons can assist the mentioned task. We use a tagged corpus (Bijankhan corpus) in Persian (Farsi) as a close language to Kurdish to develop a POS-tagged lexicon. This paper presents the approach of leveraging the resource of a close language to Kurdish to enrich its resources. A partial dataset of the results is publicly available for non-commercial use under CC BY-NC-SA 4.0 license at \url{https://kurdishblark.github.io/}. We plan to make the whole tagged corpus available after further investigation on the outcome. The dataset can help in developing POS-tagged lexicons for other Kurdish dialects and automated Kurdish corpora tagging.   

\end{abstract}

\section{Introduction}
\label{sec:intro}

The Kurdish language has received more attention since \newcite{hassani2018blark} reported the status of resource availability for the language. However, the language requires much more effort and attention to gain a position that could be called ``computable''~\cite{hassani2018blark}. As for other languages, one of the most costly areas of resources development for Kurdish processing is the preparation of tagged-corpora. The Part of Speech Tagging (POST) is essentially a manual, time-consuming, and human-resource demanding task \cite{aluisio2003account,tsai2004reliable}. Various Natural Language Processing (NLP) tasks such as Machine Translation (MT), Named Entity Recognition (NER), and Information Retrieval (IR) use POST \cite{araujo2002part}. Therefore, those tasks can benefit from automating POST even if the automation is partial. Furthermore, in multi-dialect languages such as Kurdish, having resources in one dialect can be used to develop and expand the resources for the other dialects of the language.

This paper presents the an attempt to leverage the resources of a close language to Kurdish to enrich its resources. Using resources from a close language to POS tag another has been reported in the literature~\cite{hana2006tagging,khan2011projecting,scherrer2013lexicon,vergez2013tagging,hamdi2015pos,turki2016pos,magistry2019exploiting,eskander2020unsupervised} in which different solutions have been proposed. We use Bijankhan \cite{oroumchian2006creating} corpus in Persian (Farsi) as a close language to Kurdish to develop a POS-tagged lexicon. 

The rest of this paper is organized as follows. Section \ref{sec:rw} reviews the related work. In Section \ref{sec:method}, we present the method that we follow to compute the linguistic distance. We report the results in Section~\ref{sec:result}. Finally, Section \ref{sec:conc} concludes the paper and suggests some areas for future work.

\section{Related Work}
\label{sec:rw}

The Kurdish POS tagged resources are scarce despite the efforts to improve its status by various scholars active in Kurdish processing to initiate projects. In this section, we address  Kurdish POS tagging work. 

\newcite{UoM} tagged the transcribed data of the collected speech data set from different Kurdish-speaking regions. The annotation alongside a search utility allows researchers to retrieve the data from various phonological and morphological perspectives. 

A team of scholars at the University of Kurdistan in Sanandaj developed a corpus for Kurdish (Sorani) ~\cite{UoK-corpus}. According to the project website~\cite{UoK-corpus},  the team has tagged a part of the corpus so far. The website offers a search facility to allow users to search the corpus based on the tags or specific words. 

Similarly, about 22\% AsoSoft-Corpus~\cite{AsoSoft}) for Sorani is annotated according to six categories~\cite{veisi2019toward}. The developers have normalized the text and used the Text Encoding Initiative (TEI) XML format to present the data.

KSLexicon~\cite{mohammadamini2015KSLexicon} includes 35,000 Sorani entries that have been tagged based on 28 part-of-speech tags. The developers state that the data has been extracted from an electronic Kurdish-Persian dictionary.

\newcite{walther2010developing} suggested a method for developing a lexicon for less-resourced languages and applied it to Sorani. Their morphological lexicon, SoraLex, is publically available\footnote{Soralex is available at~\url{https://gforge.inria.fr/frs/?group_id=482}}.

One of the earliest work on Kurdish POST is a ``a pre-annotation tool for developing a POS-annotated corpus'', for Kurmanji that its developers called it KurLex \cite{walther2010fast}. 

The literature has many resources that address the usage of resourceful languages in POS-tagging under-resourced languages~\cite{hana2006tagging,khan2011projecting,scherrer2013lexicon,vergez2013tagging,hamdi2015pos,turki2016pos,magistry2019exploiting,eskander2020unsupervised}. The related work have applied different approaches and methods in leveraging a resourceful language in favor of a low-resourced one. Some of those approaches have been used in POS-tagging of Kurdish in the research that we mentioned above.
   
However, the literature shows that the preparation of POS-tagged lexicon and POS-tagged corpora for Kurdish remains one of the high-priority and challenging tasks in Kurdish processing. The automation of the process could help in developing more resources in a shorter period. Although traditional POS-tagging might produce more accurate outcomes, it takes a much longer time and requires more costly efforts. Therefore, the automation approaches to Kurdish POS tagging should be intensified along with the continuation of the manual efforts. 

\section{Method}
\label{sec:method}

We use a combination of manual and automated processes to prepare a POS-tagged lexicon from a close language to Kurdish. We translate Bijankahn POS-tagged to Kurdish (Sorani). We manually check the translated entries by labeling them as ``correct'' for the entries translated correctly, ``not-correct'' for the wrong translations, and ``not-sure'' for those that we are not certain about the correctness of their translation. We then organize the result into three lists according to the mentioned labels. 

In this paper, we focus on the \textit{correct} list and leave the rest for future work. We manually process the \textit{correct} list further to check the accuracy and the relevance of the tags for the translated lexicon. If we can increase the accuracy through trivial changes, we apply the necessary revisions, or we label them as ``concerned'' otherwise. The final list includes the entries that are correct from both translation and POS-tagging perspectives.  

\begin{figure}
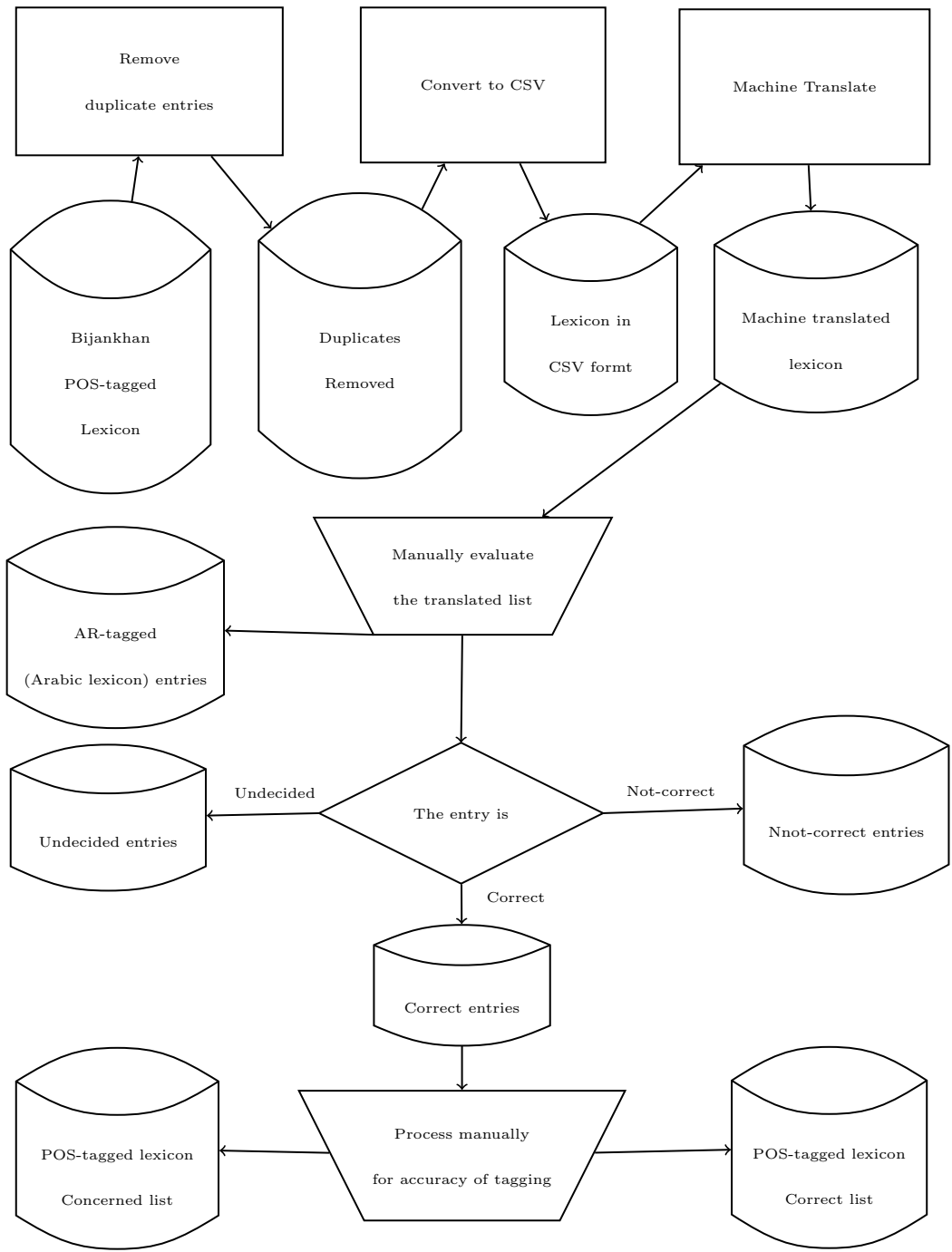

\begin{center}
	\include{POST-method}
	\caption{The process flow of the applied method}
	\label{fig:method}
\end{center}
\end{figure}

We use Microsoft Bing\footnote{To expedite the task, we also use \url{https://www.stars21.com/translator/persian/kurdish/}} for the machine translation. We also use a combination of Libreoffice calc and python scripts for other automated or semi-automated processes. Figure~\ref{fig:method} illustrates the flow of the process. 
    
\section{Result}
\label{sec:result}

Applying the method described in Section~\ref{sec:method}, we obtained several datasets that are shown in Table~\ref{tab:res}.

\begin{table}
	\begin{center}
	\caption{The input/output datasets to/from the processes}
	\label{tab:res}
	\scalebox{0.82}{
	\begin{tabular}{|c|c|c|c|c|}
		\hline
		{}&{Input dataset} & {Number of Entries} & {Output dataset} & {Number of Entries} \\ \hline \hline 
		{Remove duplicates}& {Collection UNI.txt} & {2,597,937} & {Duplicates-removed.txt} & {84,467} \\ \hline
		{Convert to CSV}& {Duplicates-removed.txt} & {84,467} & {Duplicates-removed.csv} & {84,467} \\ \hline
		{Machine translation}& {Duplicates-removed.csv} & {84,467} & {Translated.csv} & {84,467} \\ \hline
		{Evaluate the}& {Duplicates-removed.csv} & {84,467} & {Correct} & {20,059$^a$}\\ 
		{translated output}& {} & {} & {Not-correct} & {55,684} \\ 
		{}& {} & {} & {Undecided} & {8,700} \\ 
		{}& {} & {} & {AR-tagged} & {1,845} \\ \hline
		{Evaluate the}& {Correct dataset} & {20,059} & {Accurate} & {13,294}\\
		{accuracy of tagging}& {} & {} & {Repeated$^b$} & {6,494}\\ 
		{}& {} & {} & {Concerned} & {271}\\ 
		\hline
	\end{tabular}
}
	\end{center}
\footnotesize\tiny{$^a$The list had 20,083 entries. Further processes found 14 repeated Farsi entries.}\\
\footnotesize\tiny{$^b$Repeated Kurdish entries.}
\end{table}

During the manual process, we applied trivial changes such as removing avoidable {\tiny``\RL{من}''} and {\tiny``\RL{تۆ}''} at the beginning or end of some vocabularies. Furthermore, the duplicates that couldn't be identified automatically were also removed.

Table~\ref{tab:post-stat} shows the number, percentage, and percentile of each POST in the resulted lexicon. As the data shows the highest number belongs to the singular nouns (N\_SING) with 6,998 entries (about 52\% of the lexicon) and the smallest number belongs to Oh nouns (OHH) with 1 entry.

\begin{table}
	\begin{center}
		\caption{The number, percentage, and percentile of each POST in the resulted lexicon.}
		\label{tab:post-stat}
		\scalebox{1}{
		\begin{tabular}{|c|c|c|c|c|c|}
			\hline
			POS	&Lexicon No	&\%	&Percentile	\\ \hline \hline
			ADJ	&4	&0.0003	&0.1	\\ \hline
			ADJ\_CMPR	&133	&0.0099	&0.8	\\ \hline
			ADJ\_INO	&52	&0.0039	&0.6	\\ \hline
			ADJ\_ORD	&21	&0.0016	&0.4	\\ \hline
			ADJ\_SIM	&2181	&0.1629	&0.9	\\ \hline
			ADJ\_SUP	&153	&0.0114	&0.8	\\ \hline
			ADV	&23	&0.0017	&0.4	\\ \hline
			ADV\_EXM	&8	&0.0006	&0.2	\\ \hline
			ADV\_I	&11	&0.0008	&0.3	\\ \hline
			ADV\_NEGG	&8	&0.0006	&0.2	\\ \hline
			ADV\_NI	&314	&0.0235	&0.9	\\ \hline
			ADV\_TIME	&58	&0.0043	&0.7	\\ \hline
			CON	&119	&0.0089	&0.7	\\ \hline
			DEFAULT	&5	&0.0004	&0.2	\\ \hline
			DELM	&75	&0.0056	&0.7	\\ \hline
			DET	&14	&0.0010	&0.3	\\ \hline
			IF	&4	&0.0003	&0.1	\\ \hline
			INT	&4	&0.0003	&0.1	\\ \hline
			MORP	&25	&0.0019	&0.4	\\ \hline
			MQUA	&3	&0.0002	&0.08	\\ \hline
			N\_PL	&2147	&0.1604	&0.9	\\ \hline
			N\_SING	&6998	&0.5228	&1	\\ \hline
			NP	&6	&0.0004	&0.2	\\ \hline
			OH	&2	&0.0001	&0.05	\\ \hline
			OHH	&1	&0.0001	&0.03	\\ \hline
			P	&50	&0.0037	&0.6	\\ \hline
			PP	&27	&0.0020	&0.5	\\ \hline
			PRO	&44	&0.0033	&0.6	\\ \hline
			PS	&15	&0.0011	&0.3	\\ \hline
			QUA	&29	&0.0022	&0.5	\\ \hline
			SPEC	&34	&0.0025	&0.5	\\ \hline
			V\_AUX	&22	&0.0016	&0.4	\\ \hline
			V\_IMP	&48	&0.0036	&0.6	\\ \hline
			V\_PA	&274	&0.0205	&0.9	\\ \hline
			V\_PRE	&209	&0.0156	&0.8	\\ \hline
			V\_PRS	&165	&0.0123	&0.8	\\ \hline
			V\_SUB	&99	&0.0074	&0.7	\\ \hline
		\end{tabular}
	}
	\end{center}
\end{table}

\begin{figure}[ht!]
	\begin{center}
		\includegraphics[width=0.60\linewidth]{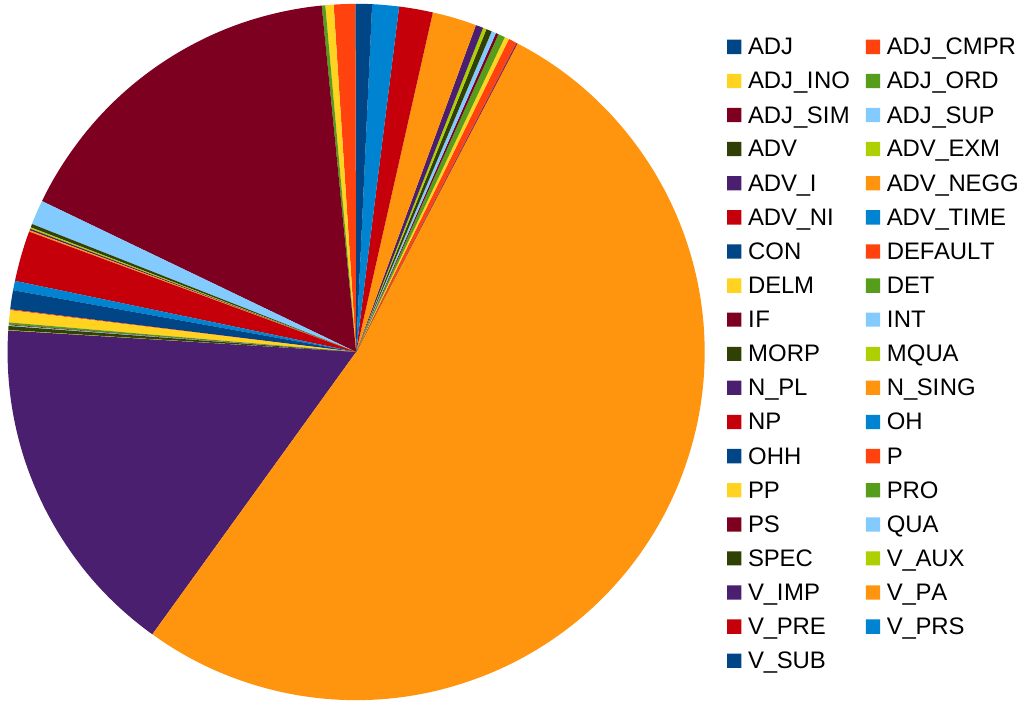}
		\caption{The magnitude POS tags in the result.}
		\label{fig:post-pie}
	\end{center}
\end{figure}

Figure~\ref{fig:post-pie} illustrates the magnitude POS tags in the result.

\begin{figure}[ht!]
	\begin{center}
		\includegraphics[width=0.60\linewidth]{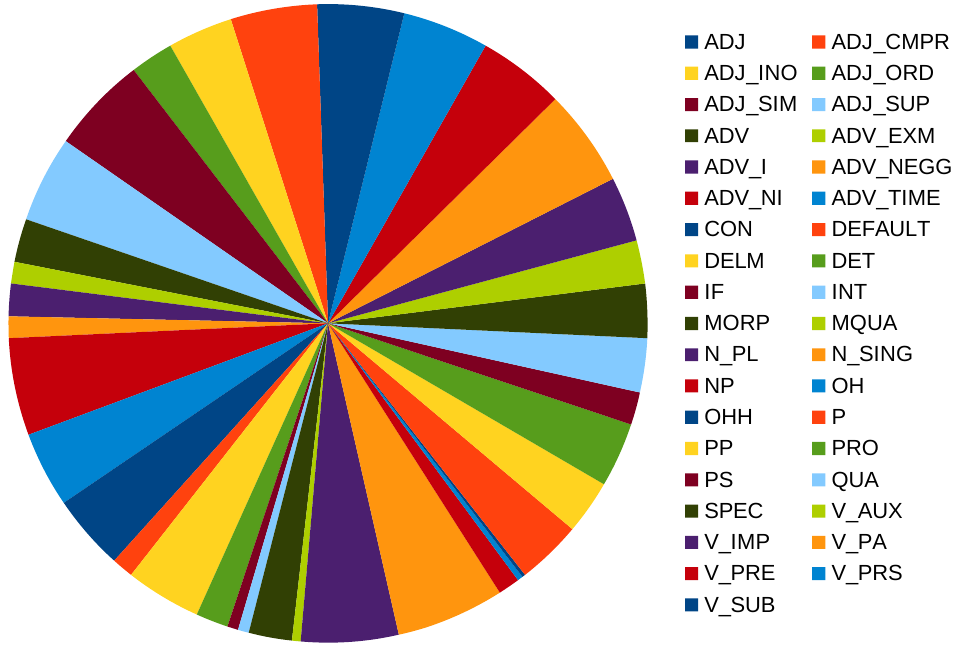}
		\caption{POS tags' percentile in the result.}
		\label{fig:post-prcntl}
	\end{center}
\end{figure}

Figure~\ref{fig:post-prcntl} presents the POS tags' percentile in the result.  

The obtained lexicon is considerably smaller than the original Farsi one. However, future work on the remaining lexicon can improve the results. That is because proper translation can eliminate the manual POS-tagging task that is the most costly activity in the preparation of the POS-tagged lexicon. The current list is also important because it allows us to use the more detailed tagset that the Bijankhan corpus suggests.

\section{Conclusion}
\label{sec:conc}
This paper provides the process and the result of obtaining a Kurdish (Sorani) POS-tagged lexicon form a POS-tagged corpus in Farsi (Persian) through combining machine translation, automated duplication removal, and manual evaluation of the results. Through the mentioned process we obtained a POS-tagged lexicon of 13,294 entries for Kurdish (Sorani). The lexicon can be used as a seed for further development that can be expanded to other Kurdish dialects. It is also important as has been tagged using a considerably wider tagset that can help in tagging Kurdish corpora more precisely.   

In the future, we would like to expand the lexicon ot other Kurdish dialects, apply it as a seed to expand its volume using synonyms, and using it to tag available Kurdish corpora.

\bibliographystyle{lrec}
\bibliography{KurdishPOST}

\end{document}

%% file: POST-method.tex
\ifx\du\undefined
  \newlength{\du}
\fi
\setlength{\du}{15\unitlength}
\begin{tikzpicture}[even odd rule]
\tiny
\pgftransformxscale{1.000000}
\pgftransformyscale{-1.000000}
\definecolor{dialinecolor}{rgb}{0.000000, 0.000000, 0.000000}
\pgfsetstrokecolor{dialinecolor}
\pgfsetstrokeopacity{1.000000}
\definecolor{diafillcolor}{rgb}{1.000000, 1.000000, 1.000000}
\pgfsetfillcolor{diafillcolor}
\pgfsetfillopacity{1.000000}
\pgfsetlinewidth{0.050000\du}
\pgfsetdash{}{0pt}
\pgfsetmiterjoin
\definecolor{diafillcolor}{rgb}{1.000000, 1.000000, 1.000000}
\pgfsetfillcolor{diafillcolor}
\pgfsetfillopacity{1.000000}
\fill (16.650010\du,23.095300\du)--(20.559420\du,25.050005\du)--(16.650010\du,27.004710\du)--(12.740600\du,25.050005\du)--cycle;
\definecolor{dialinecolor}{rgb}{0.000000, 0.000000, 0.000000}
\pgfsetstrokecolor{dialinecolor}
\pgfsetstrokeopacity{1.000000}
\draw (16.650010\du,23.095300\du)--(20.559420\du,25.050005\du)--(16.650010\du,27.004710\du)--(12.740600\du,25.050005\du)--cycle;
\definecolor{dialinecolor}{rgb}{0.000000, 0.000000, 0.000000}
\pgfsetstrokecolor{dialinecolor}
\pgfsetstrokeopacity{1.000000}
\definecolor{diafillcolor}{rgb}{0.000000, 0.000000, 0.000000}
\pgfsetfillcolor{diafillcolor}
\pgfsetfillopacity{1.000000}
\node[anchor=base,inner sep=0pt, outer sep=0pt,color=dialinecolor] at (16.650010\du,25.192505\du){The entry is};
\pgfsetlinewidth{0.050000\du}
\pgfsetdash{}{0pt}
\pgfsetbuttcap
\pgfsetmiterjoin
\pgfsetlinewidth{0.050000\du}
\pgfsetbuttcap
\pgfsetmiterjoin
\pgfsetdash{}{0pt}
\definecolor{diafillcolor}{rgb}{1.000000, 1.000000, 1.000000}
\pgfsetfillcolor{diafillcolor}
\pgfsetfillopacity{1.000000}
\definecolor{dialinecolor}{rgb}{0.000000, 0.000000, 0.000000}
\pgfsetstrokecolor{dialinecolor}
\pgfsetstrokeopacity{1.000000}
\pgfpathmoveto{\pgfpoint{24.434400\du}{23.175000\du}}
\pgfpathcurveto{\pgfpoint{25.562400\du}{22.556250\du}}{\pgfpoint{26.126400\du}{22.350000\du}}{\pgfpoint{27.254400\du}{22.350000\du}}
\pgfpathcurveto{\pgfpoint{28.382400\du}{22.350000\du}}{\pgfpoint{28.946400\du}{22.556250\du}}{\pgfpoint{30.074400\du}{23.175000\du}}
\pgfpathlineto{\pgfpoint{30.074400\du}{26.475000\du}}
\pgfpathcurveto{\pgfpoint{28.946400\du}{27.093750\du}}{\pgfpoint{28.382400\du}{27.300000\du}}{\pgfpoint{27.254400\du}{27.300000\du}}
\pgfpathcurveto{\pgfpoint{26.126400\du}{27.300000\du}}{\pgfpoint{25.562400\du}{27.093750\du}}{\pgfpoint{24.434400\du}{26.475000\du}}
\pgfpathlineto{\pgfpoint{24.434400\du}{23.175000\du}}
\pgfpathclose
\pgfusepath{fill,stroke}
\pgfsetbuttcap
\pgfsetmiterjoin
\pgfsetdash{}{0pt}
\definecolor{dialinecolor}{rgb}{0.000000, 0.000000, 0.000000}
\pgfsetstrokecolor{dialinecolor}
\pgfsetstrokeopacity{1.000000}
\pgfpathmoveto{\pgfpoint{24.434400\du}{23.175000\du}}
\pgfpathcurveto{\pgfpoint{25.562400\du}{23.793750\du}}{\pgfpoint{26.126400\du}{24.000000\du}}{\pgfpoint{27.254400\du}{24.000000\du}}
\pgfpathcurveto{\pgfpoint{28.382400\du}{24.000000\du}}{\pgfpoint{28.946400\du}{23.793750\du}}{\pgfpoint{30.074400\du}{23.175000\du}}
\pgfusepath{stroke}
\definecolor{dialinecolor}{rgb}{0.000000, 0.000000, 0.000000}
\pgfsetstrokecolor{dialinecolor}
\pgfsetstrokeopacity{1.000000}
\definecolor{diafillcolor}{rgb}{0.000000, 0.000000, 0.000000}
\pgfsetfillcolor{diafillcolor}
\pgfsetfillopacity{1.000000}
\node[anchor=base,inner sep=0pt, outer sep=0pt,color=dialinecolor] at (27.254400\du,25.078750\du){ };
\definecolor{dialinecolor}{rgb}{0.000000, 0.000000, 0.000000}
\pgfsetstrokecolor{dialinecolor}
\pgfsetstrokeopacity{1.000000}
\definecolor{diafillcolor}{rgb}{0.000000, 0.000000, 0.000000}
\pgfsetfillcolor{diafillcolor}
\pgfsetfillopacity{1.000000}
\node[anchor=base,inner sep=0pt, outer sep=0pt,color=dialinecolor] at (27.254400\du,25.713750\du){Nnot-correct entries };
\pgfsetlinewidth{0.050000\du}
\pgfsetdash{}{0pt}
\pgfsetbuttcap
\pgfsetmiterjoin
\pgfsetlinewidth{0.050000\du}
\pgfsetbuttcap
\pgfsetmiterjoin
\pgfsetdash{}{0pt}
\definecolor{diafillcolor}{rgb}{1.000000, 1.000000, 1.000000}
\pgfsetfillcolor{diafillcolor}
\pgfsetfillopacity{1.000000}
\definecolor{dialinecolor}{rgb}{0.000000, 0.000000, 0.000000}
\pgfsetstrokecolor{dialinecolor}
\pgfsetstrokeopacity{1.000000}
\pgfpathmoveto{\pgfpoint{4.250000\du}{23.825000\du}}
\pgfpathcurveto{\pgfpoint{5.324500\du}{23.318750\du}}{\pgfpoint{5.861750\du}{23.150000\du}}{\pgfpoint{6.936250\du}{23.150000\du}}
\pgfpathcurveto{\pgfpoint{8.010750\du}{23.150000\du}}{\pgfpoint{8.548000\du}{23.318750\du}}{\pgfpoint{9.622500\du}{23.825000\du}}
\pgfpathlineto{\pgfpoint{9.622500\du}{26.525000\du}}
\pgfpathcurveto{\pgfpoint{8.548000\du}{27.031250\du}}{\pgfpoint{8.010750\du}{27.200000\du}}{\pgfpoint{6.936250\du}{27.200000\du}}
\pgfpathcurveto{\pgfpoint{5.861750\du}{27.200000\du}}{\pgfpoint{5.324500\du}{27.031250\du}}{\pgfpoint{4.250000\du}{26.525000\du}}
\pgfpathlineto{\pgfpoint{4.250000\du}{23.825000\du}}
\pgfpathclose
\pgfusepath{fill,stroke}
\pgfsetbuttcap
\pgfsetmiterjoin
\pgfsetdash{}{0pt}
\definecolor{dialinecolor}{rgb}{0.000000, 0.000000, 0.000000}
\pgfsetstrokecolor{dialinecolor}
\pgfsetstrokeopacity{1.000000}
\pgfpathmoveto{\pgfpoint{4.250000\du}{23.825000\du}}
\pgfpathcurveto{\pgfpoint{5.324500\du}{24.331250\du}}{\pgfpoint{5.861750\du}{24.500000\du}}{\pgfpoint{6.936250\du}{24.500000\du}}
\pgfpathcurveto{\pgfpoint{8.010750\du}{24.500000\du}}{\pgfpoint{8.548000\du}{24.331250\du}}{\pgfpoint{9.622500\du}{23.825000\du}}
\pgfusepath{stroke}
\definecolor{dialinecolor}{rgb}{0.000000, 0.000000, 0.000000}
\pgfsetstrokecolor{dialinecolor}
\pgfsetstrokeopacity{1.000000}
\definecolor{diafillcolor}{rgb}{0.000000, 0.000000, 0.000000}
\pgfsetfillcolor{diafillcolor}
\pgfsetfillopacity{1.000000}
\node[anchor=base,inner sep=0pt, outer sep=0pt,color=dialinecolor] at (6.936250\du,25.353750\du){};
\definecolor{dialinecolor}{rgb}{0.000000, 0.000000, 0.000000}
\pgfsetstrokecolor{dialinecolor}
\pgfsetstrokeopacity{1.000000}
\definecolor{diafillcolor}{rgb}{0.000000, 0.000000, 0.000000}
\pgfsetfillcolor{diafillcolor}
\pgfsetfillopacity{1.000000}
\node[anchor=base,inner sep=0pt, outer sep=0pt,color=dialinecolor] at (6.936250\du,25.988750\du){Undecided entries};
\pgfsetlinewidth{0.050000\du}
\pgfsetdash{}{0pt}
\pgfsetbuttcap
\pgfsetmiterjoin
\pgfsetlinewidth{0.050000\du}
\pgfsetbuttcap
\pgfsetmiterjoin
\pgfsetdash{}{0pt}
\definecolor{diafillcolor}{rgb}{1.000000, 1.000000, 1.000000}
\pgfsetfillcolor{diafillcolor}
\pgfsetfillopacity{1.000000}
\definecolor{dialinecolor}{rgb}{0.000000, 0.000000, 0.000000}
\pgfsetstrokecolor{dialinecolor}
\pgfsetstrokeopacity{1.000000}
\pgfpathmoveto{\pgfpoint{4.147500\du}{18.040000\du}}
\pgfpathcurveto{\pgfpoint{5.342500\du}{17.342500\du}}{\pgfpoint{5.940000\du}{17.110000\du}}{\pgfpoint{7.135000\du}{17.110000\du}}
\pgfpathcurveto{\pgfpoint{8.330000\du}{17.110000\du}}{\pgfpoint{8.927500\du}{17.342500\du}}{\pgfpoint{10.122500\du}{18.040000\du}}
\pgfpathlineto{\pgfpoint{10.122500\du}{21.760000\du}}
\pgfpathcurveto{\pgfpoint{8.927500\du}{22.457500\du}}{\pgfpoint{8.330000\du}{22.690000\du}}{\pgfpoint{7.135000\du}{22.690000\du}}
\pgfpathcurveto{\pgfpoint{5.940000\du}{22.690000\du}}{\pgfpoint{5.342500\du}{22.457500\du}}{\pgfpoint{4.147500\du}{21.760000\du}}
\pgfpathlineto{\pgfpoint{4.147500\du}{18.040000\du}}
\pgfpathclose
\pgfusepath{fill,stroke}
\pgfsetbuttcap
\pgfsetmiterjoin
\pgfsetdash{}{0pt}
\definecolor{dialinecolor}{rgb}{0.000000, 0.000000, 0.000000}
\pgfsetstrokecolor{dialinecolor}
\pgfsetstrokeopacity{1.000000}
\pgfpathmoveto{\pgfpoint{4.147500\du}{18.040000\du}}
\pgfpathcurveto{\pgfpoint{5.342500\du}{18.737500\du}}{\pgfpoint{5.940000\du}{18.970000\du}}{\pgfpoint{7.135000\du}{18.970000\du}}
\pgfpathcurveto{\pgfpoint{8.330000\du}{18.970000\du}}{\pgfpoint{8.927500\du}{18.737500\du}}{\pgfpoint{10.122500\du}{18.040000\du}}
\pgfusepath{stroke}
\definecolor{dialinecolor}{rgb}{0.000000, 0.000000, 0.000000}
\pgfsetstrokecolor{dialinecolor}
\pgfsetstrokeopacity{1.000000}
\definecolor{diafillcolor}{rgb}{0.000000, 0.000000, 0.000000}
\pgfsetfillcolor{diafillcolor}
\pgfsetfillopacity{1.000000}
\node[anchor=base,inner sep=0pt, outer sep=0pt,color=dialinecolor] at (7.135000\du,19.571250\du){};
\definecolor{dialinecolor}{rgb}{0.000000, 0.000000, 0.000000}
\pgfsetstrokecolor{dialinecolor}
\pgfsetstrokeopacity{1.000000}
\definecolor{diafillcolor}{rgb}{0.000000, 0.000000, 0.000000}
\pgfsetfillcolor{diafillcolor}
\pgfsetfillopacity{1.000000}
\node[anchor=base,inner sep=0pt, outer sep=0pt,color=dialinecolor] at (7.135000\du,20.206250\du){AR-tagged};
\definecolor{dialinecolor}{rgb}{0.000000, 0.000000, 0.000000}
\pgfsetstrokecolor{dialinecolor}
\pgfsetstrokeopacity{1.000000}
\definecolor{diafillcolor}{rgb}{0.000000, 0.000000, 0.000000}
\pgfsetfillcolor{diafillcolor}
\pgfsetfillopacity{1.000000}
\node[anchor=base,inner sep=0pt, outer sep=0pt,color=dialinecolor] at (7.135000\du,20.841250\du){};
\definecolor{dialinecolor}{rgb}{0.000000, 0.000000, 0.000000}
\pgfsetstrokecolor{dialinecolor}
\pgfsetstrokeopacity{1.000000}
\definecolor{diafillcolor}{rgb}{0.000000, 0.000000, 0.000000}
\pgfsetfillcolor{diafillcolor}
\pgfsetfillopacity{1.000000}
\node[anchor=base,inner sep=0pt, outer sep=0pt,color=dialinecolor] at (7.135000\du,21.476250\du){ (Arabic lexicon) entries};
\pgfsetlinewidth{0.050000\du}
\pgfsetdash{}{0pt}
\pgfsetbuttcap
\pgfsetmiterjoin
\pgfsetlinewidth{0.050000\du}
\pgfsetbuttcap
\pgfsetmiterjoin
\pgfsetdash{}{0pt}
\definecolor{diafillcolor}{rgb}{1.000000, 1.000000, 1.000000}
\pgfsetfillcolor{diafillcolor}
\pgfsetfillopacity{1.000000}
\definecolor{dialinecolor}{rgb}{0.000000, 0.000000, 0.000000}
\pgfsetstrokecolor{dialinecolor}
\pgfsetstrokeopacity{1.000000}
\pgfpathmoveto{\pgfpoint{14.250000\du}{28.708333\du}}
\pgfpathcurveto{\pgfpoint{15.220000\du}{28.289583\du}}{\pgfpoint{15.705000\du}{28.150000\du}}{\pgfpoint{16.675000\du}{28.150000\du}}
\pgfpathcurveto{\pgfpoint{17.645000\du}{28.150000\du}}{\pgfpoint{18.130000\du}{28.289583\du}}{\pgfpoint{19.100000\du}{28.708333\du}}
\pgfpathlineto{\pgfpoint{19.100000\du}{30.941667\du}}
\pgfpathcurveto{\pgfpoint{18.130000\du}{31.360417\du}}{\pgfpoint{17.645000\du}{31.500000\du}}{\pgfpoint{16.675000\du}{31.500000\du}}
\pgfpathcurveto{\pgfpoint{15.705000\du}{31.500000\du}}{\pgfpoint{15.220000\du}{31.360417\du}}{\pgfpoint{14.250000\du}{30.941667\du}}
\pgfpathlineto{\pgfpoint{14.250000\du}{28.708333\du}}
\pgfpathclose
\pgfusepath{fill,stroke}
\pgfsetbuttcap
\pgfsetmiterjoin
\pgfsetdash{}{0pt}
\definecolor{dialinecolor}{rgb}{0.000000, 0.000000, 0.000000}
\pgfsetstrokecolor{dialinecolor}
\pgfsetstrokeopacity{1.000000}
\pgfpathmoveto{\pgfpoint{14.250000\du}{28.708333\du}}
\pgfpathcurveto{\pgfpoint{15.220000\du}{29.127083\du}}{\pgfpoint{15.705000\du}{29.266667\du}}{\pgfpoint{16.675000\du}{29.266667\du}}
\pgfpathcurveto{\pgfpoint{17.645000\du}{29.266667\du}}{\pgfpoint{18.130000\du}{29.127083\du}}{\pgfpoint{19.100000\du}{28.708333\du}}
\pgfusepath{stroke}
\definecolor{dialinecolor}{rgb}{0.000000, 0.000000, 0.000000}
\pgfsetstrokecolor{dialinecolor}
\pgfsetstrokeopacity{1.000000}
\definecolor{diafillcolor}{rgb}{0.000000, 0.000000, 0.000000}
\pgfsetfillcolor{diafillcolor}
\pgfsetfillopacity{1.000000}
\node[anchor=base,inner sep=0pt, outer sep=0pt,color=dialinecolor] at (16.675000\du,29.945417\du){};
\definecolor{dialinecolor}{rgb}{0.000000, 0.000000, 0.000000}
\pgfsetstrokecolor{dialinecolor}
\pgfsetstrokeopacity{1.000000}
\definecolor{diafillcolor}{rgb}{0.000000, 0.000000, 0.000000}
\pgfsetfillcolor{diafillcolor}
\pgfsetfillopacity{1.000000}
\node[anchor=base,inner sep=0pt, outer sep=0pt,color=dialinecolor] at (16.675000\du,30.580417\du){Correct entries};
\pgfsetlinewidth{0.050000\du}
\pgfsetdash{}{0pt}
\pgfsetbuttcap
\pgfsetmiterjoin
\pgfsetlinewidth{0.050000\du}
\pgfsetbuttcap
\pgfsetmiterjoin
\pgfsetdash{}{0pt}
\definecolor{diafillcolor}{rgb}{1.000000, 1.000000, 1.000000}
\pgfsetfillcolor{diafillcolor}
\pgfsetfillopacity{1.000000}
\definecolor{dialinecolor}{rgb}{0.000000, 0.000000, 0.000000}
\pgfsetstrokecolor{dialinecolor}
\pgfsetstrokeopacity{1.000000}
\pgfpathmoveto{\pgfpoint{4.250000\du}{9.411673\du}}
\pgfpathcurveto{\pgfpoint{5.351000\du}{8.396673\du}}{\pgfpoint{5.901500\du}{8.058340\du}}{\pgfpoint{7.002500\du}{8.058340\du}}
\pgfpathcurveto{\pgfpoint{8.103500\du}{8.058340\du}}{\pgfpoint{8.654000\du}{8.396673\du}}{\pgfpoint{9.755000\du}{9.411673\du}}
\pgfpathlineto{\pgfpoint{9.755000\du}{14.825006\du}}
\pgfpathcurveto{\pgfpoint{8.654000\du}{15.840006\du}}{\pgfpoint{8.103500\du}{16.178340\du}}{\pgfpoint{7.002500\du}{16.178340\du}}
\pgfpathcurveto{\pgfpoint{5.901500\du}{16.178340\du}}{\pgfpoint{5.351000\du}{15.840006\du}}{\pgfpoint{4.250000\du}{14.825006\du}}
\pgfpathlineto{\pgfpoint{4.250000\du}{9.411673\du}}
\pgfpathclose
\pgfusepath{fill,stroke}
\pgfsetbuttcap
\pgfsetmiterjoin
\pgfsetdash{}{0pt}
\definecolor{dialinecolor}{rgb}{0.000000, 0.000000, 0.000000}
\pgfsetstrokecolor{dialinecolor}
\pgfsetstrokeopacity{1.000000}
\pgfpathmoveto{\pgfpoint{4.250000\du}{9.411673\du}}
\pgfpathcurveto{\pgfpoint{5.351000\du}{10.426673\du}}{\pgfpoint{5.901500\du}{10.765007\du}}{\pgfpoint{7.002500\du}{10.765007\du}}
\pgfpathcurveto{\pgfpoint{8.103500\du}{10.765007\du}}{\pgfpoint{8.654000\du}{10.426673\du}}{\pgfpoint{9.755000\du}{9.411673\du}}
\pgfusepath{stroke}
\definecolor{dialinecolor}{rgb}{0.000000, 0.000000, 0.000000}
\pgfsetstrokecolor{dialinecolor}
\pgfsetstrokeopacity{1.000000}
\definecolor{diafillcolor}{rgb}{0.000000, 0.000000, 0.000000}
\pgfsetfillcolor{diafillcolor}
\pgfsetfillopacity{1.000000}
\node[anchor=base,inner sep=0pt, outer sep=0pt,color=dialinecolor] at (7.002500\du,11.366257\du){};
\definecolor{dialinecolor}{rgb}{0.000000, 0.000000, 0.000000}
\pgfsetstrokecolor{dialinecolor}
\pgfsetstrokeopacity{1.000000}
\definecolor{diafillcolor}{rgb}{0.000000, 0.000000, 0.000000}
\pgfsetfillcolor{diafillcolor}
\pgfsetfillopacity{1.000000}
\node[anchor=base,inner sep=0pt, outer sep=0pt,color=dialinecolor] at (7.002500\du,12.001257\du){Bijankhan};
\definecolor{dialinecolor}{rgb}{0.000000, 0.000000, 0.000000}
\pgfsetstrokecolor{dialinecolor}
\pgfsetstrokeopacity{1.000000}
\definecolor{diafillcolor}{rgb}{0.000000, 0.000000, 0.000000}
\pgfsetfillcolor{diafillcolor}
\pgfsetfillopacity{1.000000}
\node[anchor=base,inner sep=0pt, outer sep=0pt,color=dialinecolor] at (7.002500\du,12.636257\du){ };
\definecolor{dialinecolor}{rgb}{0.000000, 0.000000, 0.000000}
\pgfsetstrokecolor{dialinecolor}
\pgfsetstrokeopacity{1.000000}
\definecolor{diafillcolor}{rgb}{0.000000, 0.000000, 0.000000}
\pgfsetfillcolor{diafillcolor}
\pgfsetfillopacity{1.000000}
\node[anchor=base,inner sep=0pt, outer sep=0pt,color=dialinecolor] at (7.002500\du,13.271256\du){POS-tagged};
\definecolor{dialinecolor}{rgb}{0.000000, 0.000000, 0.000000}
\pgfsetstrokecolor{dialinecolor}
\pgfsetstrokeopacity{1.000000}
\definecolor{diafillcolor}{rgb}{0.000000, 0.000000, 0.000000}
\pgfsetfillcolor{diafillcolor}
\pgfsetfillopacity{1.000000}
\node[anchor=base,inner sep=0pt, outer sep=0pt,color=dialinecolor] at (7.002500\du,13.906256\du){};
\definecolor{dialinecolor}{rgb}{0.000000, 0.000000, 0.000000}
\pgfsetstrokecolor{dialinecolor}
\pgfsetstrokeopacity{1.000000}
\definecolor{diafillcolor}{rgb}{0.000000, 0.000000, 0.000000}
\pgfsetfillcolor{diafillcolor}
\pgfsetfillopacity{1.000000}
\node[anchor=base,inner sep=0pt, outer sep=0pt,color=dialinecolor] at (7.002500\du,14.541256\du){Lexicon};
\pgfsetlinewidth{0.050000\du}
\pgfsetdash{}{0pt}
\pgfsetmiterjoin
{\pgfsetcornersarced{\pgfpoint{0.000000\du}{0.000000\du}}\definecolor{diafillcolor}{rgb}{1.000000, 1.000000, 1.000000}
\pgfsetfillcolor{diafillcolor}
\pgfsetfillopacity{1.000000}
\fill (22.662500\du,2.750000\du)--(22.662500\du,7.050000\du)--(29.547500\du,7.050000\du)--(29.547500\du,2.750000\du)--cycle;
}{\pgfsetcornersarced{\pgfpoint{0.000000\du}{0.000000\du}}\definecolor{dialinecolor}{rgb}{0.000000, 0.000000, 0.000000}
\pgfsetstrokecolor{dialinecolor}
\pgfsetstrokeopacity{1.000000}
\draw (22.662500\du,2.750000\du)--(22.662500\du,7.050000\du)--(29.547500\du,7.050000\du)--(29.547500\du,2.750000\du)--cycle;
}
\definecolor{dialinecolor}{rgb}{0.000000, 0.000000, 0.000000}
\pgfsetstrokecolor{dialinecolor}
\pgfsetstrokeopacity{1.000000}
\definecolor{diafillcolor}{rgb}{0.000000, 0.000000, 0.000000}
\pgfsetfillcolor{diafillcolor}
\pgfsetfillopacity{1.000000}
\node[anchor=base,inner sep=0pt, outer sep=0pt,color=dialinecolor] at (26.105000\du,5.042500\du){Machine Translate};
\pgfsetlinewidth{0.050000\du}
\pgfsetdash{}{0pt}
\pgfsetmiterjoin
{\pgfsetcornersarced{\pgfpoint{0.000000\du}{0.000000\du}}\definecolor{diafillcolor}{rgb}{1.000000, 1.000000, 1.000000}
\pgfsetfillcolor{diafillcolor}
\pgfsetfillopacity{1.000000}
\fill (4.400000\du,2.700000\du)--(4.400000\du,6.800000\du)--(11.740000\du,6.800000\du)--(11.740000\du,2.700000\du)--cycle;
}{\pgfsetcornersarced{\pgfpoint{0.000000\du}{0.000000\du}}\definecolor{dialinecolor}{rgb}{0.000000, 0.000000, 0.000000}
\pgfsetstrokecolor{dialinecolor}
\pgfsetstrokeopacity{1.000000}
\draw (4.400000\du,2.700000\du)--(4.400000\du,6.800000\du)--(11.740000\du,6.800000\du)--(11.740000\du,2.700000\du)--cycle;
}
\definecolor{dialinecolor}{rgb}{0.000000, 0.000000, 0.000000}
\pgfsetstrokecolor{dialinecolor}
\pgfsetstrokeopacity{1.000000}
\definecolor{diafillcolor}{rgb}{0.000000, 0.000000, 0.000000}
\pgfsetfillcolor{diafillcolor}
\pgfsetfillopacity{1.000000}
\node[anchor=base,inner sep=0pt, outer sep=0pt,color=dialinecolor] at (8.070000\du,4.257500\du){Remove };
\definecolor{dialinecolor}{rgb}{0.000000, 0.000000, 0.000000}
\pgfsetstrokecolor{dialinecolor}
\pgfsetstrokeopacity{1.000000}
\definecolor{diafillcolor}{rgb}{0.000000, 0.000000, 0.000000}
\pgfsetfillcolor{diafillcolor}
\pgfsetfillopacity{1.000000}
\node[anchor=base,inner sep=0pt, outer sep=0pt,color=dialinecolor] at (8.070000\du,4.892500\du){};
\definecolor{dialinecolor}{rgb}{0.000000, 0.000000, 0.000000}
\pgfsetstrokecolor{dialinecolor}
\pgfsetstrokeopacity{1.000000}
\definecolor{diafillcolor}{rgb}{0.000000, 0.000000, 0.000000}
\pgfsetfillcolor{diafillcolor}
\pgfsetfillopacity{1.000000}
\node[anchor=base,inner sep=0pt, outer sep=0pt,color=dialinecolor] at (8.070000\du,5.527500\du){duplicate entries};
\pgfsetlinewidth{0.050000\du}
\pgfsetdash{}{0pt}
\pgfsetbuttcap
\pgfsetmiterjoin
\pgfsetlinewidth{0.050000\du}
\pgfsetbuttcap
\pgfsetmiterjoin
\pgfsetdash{}{0pt}
\definecolor{diafillcolor}{rgb}{1.000000, 1.000000, 1.000000}
\pgfsetfillcolor{diafillcolor}
\pgfsetfillopacity{1.000000}
\definecolor{dialinecolor}{rgb}{0.000000, 0.000000, 0.000000}
\pgfsetstrokecolor{dialinecolor}
\pgfsetstrokeopacity{1.000000}
\pgfpathmoveto{\pgfpoint{17.844100\du}{9.358340\du}}
\pgfpathcurveto{\pgfpoint{18.795273\du}{8.660840\du}}{\pgfpoint{19.270860\du}{8.428340\du}}{\pgfpoint{20.222033\du}{8.428340\du}}
\pgfpathcurveto{\pgfpoint{21.173207\du}{8.428340\du}}{\pgfpoint{21.648793\du}{8.660840\du}}{\pgfpoint{22.599967\du}{9.358340\du}}
\pgfpathlineto{\pgfpoint{22.599967\du}{13.078340\du}}
\pgfpathcurveto{\pgfpoint{21.648793\du}{13.775840\du}}{\pgfpoint{21.173207\du}{14.008340\du}}{\pgfpoint{20.222033\du}{14.008340\du}}
\pgfpathcurveto{\pgfpoint{19.270860\du}{14.008340\du}}{\pgfpoint{18.795273\du}{13.775840\du}}{\pgfpoint{17.844100\du}{13.078340\du}}
\pgfpathlineto{\pgfpoint{17.844100\du}{9.358340\du}}
\pgfpathclose
\pgfusepath{fill,stroke}
\pgfsetbuttcap
\pgfsetmiterjoin
\pgfsetdash{}{0pt}
\definecolor{dialinecolor}{rgb}{0.000000, 0.000000, 0.000000}
\pgfsetstrokecolor{dialinecolor}
\pgfsetstrokeopacity{1.000000}
\pgfpathmoveto{\pgfpoint{17.844100\du}{9.358340\du}}
\pgfpathcurveto{\pgfpoint{18.795273\du}{10.055840\du}}{\pgfpoint{19.270860\du}{10.288340\du}}{\pgfpoint{20.222033\du}{10.288340\du}}
\pgfpathcurveto{\pgfpoint{21.173207\du}{10.288340\du}}{\pgfpoint{21.648793\du}{10.055840\du}}{\pgfpoint{22.599967\du}{9.358340\du}}
\pgfusepath{stroke}
\definecolor{dialinecolor}{rgb}{0.000000, 0.000000, 0.000000}
\pgfsetstrokecolor{dialinecolor}
\pgfsetstrokeopacity{1.000000}
\definecolor{diafillcolor}{rgb}{0.000000, 0.000000, 0.000000}
\pgfsetfillcolor{diafillcolor}
\pgfsetfillopacity{1.000000}
\node[anchor=base,inner sep=0pt, outer sep=0pt,color=dialinecolor] at (20.222033\du,10.889590\du){};
\definecolor{dialinecolor}{rgb}{0.000000, 0.000000, 0.000000}
\pgfsetstrokecolor{dialinecolor}
\pgfsetstrokeopacity{1.000000}
\definecolor{diafillcolor}{rgb}{0.000000, 0.000000, 0.000000}
\pgfsetfillcolor{diafillcolor}
\pgfsetfillopacity{1.000000}
\node[anchor=base,inner sep=0pt, outer sep=0pt,color=dialinecolor] at (20.222033\du,11.524590\du){Lexicon in };
\definecolor{dialinecolor}{rgb}{0.000000, 0.000000, 0.000000}
\pgfsetstrokecolor{dialinecolor}
\pgfsetstrokeopacity{1.000000}
\definecolor{diafillcolor}{rgb}{0.000000, 0.000000, 0.000000}
\pgfsetfillcolor{diafillcolor}
\pgfsetfillopacity{1.000000}
\node[anchor=base,inner sep=0pt, outer sep=0pt,color=dialinecolor] at (20.222033\du,12.159590\du){};
\definecolor{dialinecolor}{rgb}{0.000000, 0.000000, 0.000000}
\pgfsetstrokecolor{dialinecolor}
\pgfsetstrokeopacity{1.000000}
\definecolor{diafillcolor}{rgb}{0.000000, 0.000000, 0.000000}
\pgfsetfillcolor{diafillcolor}
\pgfsetfillopacity{1.000000}
\node[anchor=base,inner sep=0pt, outer sep=0pt,color=dialinecolor] at (20.222033\du,12.794590\du){CSV formt};
\pgfsetlinewidth{0.050000\du}
\pgfsetdash{}{0pt}
\pgfsetmiterjoin
{\pgfsetcornersarced{\pgfpoint{0.000000\du}{0.000000\du}}\definecolor{diafillcolor}{rgb}{1.000000, 1.000000, 1.000000}
\pgfsetfillcolor{diafillcolor}
\pgfsetfillopacity{1.000000}
\fill (13.887500\du,2.700000\du)--(13.887500\du,7.000000\du)--(20.622500\du,7.000000\du)--(20.622500\du,2.700000\du)--cycle;
}{\pgfsetcornersarced{\pgfpoint{0.000000\du}{0.000000\du}}\definecolor{dialinecolor}{rgb}{0.000000, 0.000000, 0.000000}
\pgfsetstrokecolor{dialinecolor}
\pgfsetstrokeopacity{1.000000}
\draw (13.887500\du,2.700000\du)--(13.887500\du,7.000000\du)--(20.622500\du,7.000000\du)--(20.622500\du,2.700000\du)--cycle;
}
\definecolor{dialinecolor}{rgb}{0.000000, 0.000000, 0.000000}
\pgfsetstrokecolor{dialinecolor}
\pgfsetstrokeopacity{1.000000}
\definecolor{diafillcolor}{rgb}{0.000000, 0.000000, 0.000000}
\pgfsetfillcolor{diafillcolor}
\pgfsetfillopacity{1.000000}
\node[anchor=base,inner sep=0pt, outer sep=0pt,color=dialinecolor] at (17.255000\du,4.992500\du){Convert to CSV};
\pgfsetlinewidth{0.050000\du}
\pgfsetdash{}{0pt}
\pgfsetbuttcap
\pgfsetmiterjoin
\pgfsetlinewidth{0.050000\du}
\pgfsetbuttcap
\pgfsetmiterjoin
\pgfsetdash{}{0pt}
\definecolor{diafillcolor}{rgb}{1.000000, 1.000000, 1.000000}
\pgfsetfillcolor{diafillcolor}
\pgfsetfillopacity{1.000000}
\definecolor{dialinecolor}{rgb}{0.000000, 0.000000, 0.000000}
\pgfsetstrokecolor{dialinecolor}
\pgfsetstrokeopacity{1.000000}
\pgfpathmoveto{\pgfpoint{12.185400\du}{32.750000\du}}
\pgfpathlineto{\pgfpoint{21.164567\du}{32.750000\du}}
\pgfpathlineto{\pgfpoint{19.368733\du}{36.350000\du}}
\pgfpathlineto{\pgfpoint{13.981233\du}{36.350000\du}}
\pgfpathlineto{\pgfpoint{12.185400\du}{32.750000\du}}
\pgfpathclose
\pgfusepath{fill,stroke}
\definecolor{dialinecolor}{rgb}{0.000000, 0.000000, 0.000000}
\pgfsetstrokecolor{dialinecolor}
\pgfsetstrokeopacity{1.000000}
\definecolor{diafillcolor}{rgb}{0.000000, 0.000000, 0.000000}
\pgfsetfillcolor{diafillcolor}
\pgfsetfillopacity{1.000000}
\node[anchor=base,inner sep=0pt, outer sep=0pt,color=dialinecolor] at (16.674983\du,34.073750\du){Process manually};
\definecolor{dialinecolor}{rgb}{0.000000, 0.000000, 0.000000}
\pgfsetstrokecolor{dialinecolor}
\pgfsetstrokeopacity{1.000000}
\definecolor{diafillcolor}{rgb}{0.000000, 0.000000, 0.000000}
\pgfsetfillcolor{diafillcolor}
\pgfsetfillopacity{1.000000}
\node[anchor=base,inner sep=0pt, outer sep=0pt,color=dialinecolor] at (16.674983\du,34.708750\du){};
\definecolor{dialinecolor}{rgb}{0.000000, 0.000000, 0.000000}
\pgfsetstrokecolor{dialinecolor}
\pgfsetstrokeopacity{1.000000}
\definecolor{diafillcolor}{rgb}{0.000000, 0.000000, 0.000000}
\pgfsetfillcolor{diafillcolor}
\pgfsetfillopacity{1.000000}
\node[anchor=base,inner sep=0pt, outer sep=0pt,color=dialinecolor] at (16.674983\du,35.343750\du){for accuracy of tagging};
\pgfsetlinewidth{0.050000\du}
\pgfsetdash{}{0pt}
\pgfsetbuttcap
\pgfsetmiterjoin
\pgfsetlinewidth{0.050000\du}
\pgfsetbuttcap
\pgfsetmiterjoin
\pgfsetdash{}{0pt}
\definecolor{diafillcolor}{rgb}{1.000000, 1.000000, 1.000000}
\pgfsetfillcolor{diafillcolor}
\pgfsetfillopacity{1.000000}
\definecolor{dialinecolor}{rgb}{0.000000, 0.000000, 0.000000}
\pgfsetstrokecolor{dialinecolor}
\pgfsetstrokeopacity{1.000000}
\pgfpathmoveto{\pgfpoint{24.100000\du}{32.465000\du}}
\pgfpathcurveto{\pgfpoint{25.174000\du}{31.767500\du}}{\pgfpoint{25.711000\du}{31.535000\du}}{\pgfpoint{26.785000\du}{31.535000\du}}
\pgfpathcurveto{\pgfpoint{27.859000\du}{31.535000\du}}{\pgfpoint{28.396000\du}{31.767500\du}}{\pgfpoint{29.470000\du}{32.465000\du}}
\pgfpathlineto{\pgfpoint{29.470000\du}{36.185000\du}}
\pgfpathcurveto{\pgfpoint{28.396000\du}{36.882500\du}}{\pgfpoint{27.859000\du}{37.115000\du}}{\pgfpoint{26.785000\du}{37.115000\du}}
\pgfpathcurveto{\pgfpoint{25.711000\du}{37.115000\du}}{\pgfpoint{25.174000\du}{36.882500\du}}{\pgfpoint{24.100000\du}{36.185000\du}}
\pgfpathlineto{\pgfpoint{24.100000\du}{32.465000\du}}
\pgfpathclose
\pgfusepath{fill,stroke}
\pgfsetbuttcap
\pgfsetmiterjoin
\pgfsetdash{}{0pt}
\definecolor{dialinecolor}{rgb}{0.000000, 0.000000, 0.000000}
\pgfsetstrokecolor{dialinecolor}
\pgfsetstrokeopacity{1.000000}
\pgfpathmoveto{\pgfpoint{24.100000\du}{32.465000\du}}
\pgfpathcurveto{\pgfpoint{25.174000\du}{33.162500\du}}{\pgfpoint{25.711000\du}{33.395000\du}}{\pgfpoint{26.785000\du}{33.395000\du}}
\pgfpathcurveto{\pgfpoint{27.859000\du}{33.395000\du}}{\pgfpoint{28.396000\du}{33.162500\du}}{\pgfpoint{29.470000\du}{32.465000\du}}
\pgfusepath{stroke}
\definecolor{dialinecolor}{rgb}{0.000000, 0.000000, 0.000000}
\pgfsetstrokecolor{dialinecolor}
\pgfsetstrokeopacity{1.000000}
\definecolor{diafillcolor}{rgb}{0.000000, 0.000000, 0.000000}
\pgfsetfillcolor{diafillcolor}
\pgfsetfillopacity{1.000000}
\node[anchor=base,inner sep=0pt, outer sep=0pt,color=dialinecolor] at (26.785000\du,33.996250\du){};
\definecolor{dialinecolor}{rgb}{0.000000, 0.000000, 0.000000}
\pgfsetstrokecolor{dialinecolor}
\pgfsetstrokeopacity{1.000000}
\definecolor{diafillcolor}{rgb}{0.000000, 0.000000, 0.000000}
\pgfsetfillcolor{diafillcolor}
\pgfsetfillopacity{1.000000}
\node[anchor=base,inner sep=0pt, outer sep=0pt,color=dialinecolor] at (26.785000\du,34.631250\du){POS-tagged lexicon};
\definecolor{dialinecolor}{rgb}{0.000000, 0.000000, 0.000000}
\pgfsetstrokecolor{dialinecolor}
\pgfsetstrokeopacity{1.000000}
\definecolor{diafillcolor}{rgb}{0.000000, 0.000000, 0.000000}
\pgfsetfillcolor{diafillcolor}
\pgfsetfillopacity{1.000000}
\node[anchor=base,inner sep=0pt, outer sep=0pt,color=dialinecolor] at (26.785000\du,35.266250\du){};
\definecolor{dialinecolor}{rgb}{0.000000, 0.000000, 0.000000}
\pgfsetstrokecolor{dialinecolor}
\pgfsetstrokeopacity{1.000000}
\definecolor{diafillcolor}{rgb}{0.000000, 0.000000, 0.000000}
\pgfsetfillcolor{diafillcolor}
\pgfsetfillopacity{1.000000}
\node[anchor=base,inner sep=0pt, outer sep=0pt,color=dialinecolor] at (26.785000\du,35.901250\du){Correct list};
\pgfsetlinewidth{0.050000\du}
\pgfsetdash{}{0pt}
\pgfsetbuttcap
\pgfsetmiterjoin
\pgfsetlinewidth{0.050000\du}
\pgfsetbuttcap
\pgfsetmiterjoin
\pgfsetdash{}{0pt}
\definecolor{diafillcolor}{rgb}{1.000000, 1.000000, 1.000000}
\pgfsetfillcolor{diafillcolor}
\pgfsetfillopacity{1.000000}
\definecolor{dialinecolor}{rgb}{0.000000, 0.000000, 0.000000}
\pgfsetstrokecolor{dialinecolor}
\pgfsetstrokeopacity{1.000000}
\pgfpathmoveto{\pgfpoint{4.400000\du}{32.490000\du}}
\pgfpathcurveto{\pgfpoint{5.514500\du}{31.792500\du}}{\pgfpoint{6.071750\du}{31.560000\du}}{\pgfpoint{7.186250\du}{31.560000\du}}
\pgfpathcurveto{\pgfpoint{8.300750\du}{31.560000\du}}{\pgfpoint{8.858000\du}{31.792500\du}}{\pgfpoint{9.972500\du}{32.490000\du}}
\pgfpathlineto{\pgfpoint{9.972500\du}{36.210000\du}}
\pgfpathcurveto{\pgfpoint{8.858000\du}{36.907500\du}}{\pgfpoint{8.300750\du}{37.140000\du}}{\pgfpoint{7.186250\du}{37.140000\du}}
\pgfpathcurveto{\pgfpoint{6.071750\du}{37.140000\du}}{\pgfpoint{5.514500\du}{36.907500\du}}{\pgfpoint{4.400000\du}{36.210000\du}}
\pgfpathlineto{\pgfpoint{4.400000\du}{32.490000\du}}
\pgfpathclose
\pgfusepath{fill,stroke}
\pgfsetbuttcap
\pgfsetmiterjoin
\pgfsetdash{}{0pt}
\definecolor{dialinecolor}{rgb}{0.000000, 0.000000, 0.000000}
\pgfsetstrokecolor{dialinecolor}
\pgfsetstrokeopacity{1.000000}
\pgfpathmoveto{\pgfpoint{4.400000\du}{32.490000\du}}
\pgfpathcurveto{\pgfpoint{5.514500\du}{33.187500\du}}{\pgfpoint{6.071750\du}{33.420000\du}}{\pgfpoint{7.186250\du}{33.420000\du}}
\pgfpathcurveto{\pgfpoint{8.300750\du}{33.420000\du}}{\pgfpoint{8.858000\du}{33.187500\du}}{\pgfpoint{9.972500\du}{32.490000\du}}
\pgfusepath{stroke}
\definecolor{dialinecolor}{rgb}{0.000000, 0.000000, 0.000000}
\pgfsetstrokecolor{dialinecolor}
\pgfsetstrokeopacity{1.000000}
\definecolor{diafillcolor}{rgb}{0.000000, 0.000000, 0.000000}
\pgfsetfillcolor{diafillcolor}
\pgfsetfillopacity{1.000000}
\node[anchor=base,inner sep=0pt, outer sep=0pt,color=dialinecolor] at (7.186250\du,34.021250\du){};
\definecolor{dialinecolor}{rgb}{0.000000, 0.000000, 0.000000}
\pgfsetstrokecolor{dialinecolor}
\pgfsetstrokeopacity{1.000000}
\definecolor{diafillcolor}{rgb}{0.000000, 0.000000, 0.000000}
\pgfsetfillcolor{diafillcolor}
\pgfsetfillopacity{1.000000}
\node[anchor=base,inner sep=0pt, outer sep=0pt,color=dialinecolor] at (7.186250\du,34.656250\du){POS-tagged lexicon};
\definecolor{dialinecolor}{rgb}{0.000000, 0.000000, 0.000000}
\pgfsetstrokecolor{dialinecolor}
\pgfsetstrokeopacity{1.000000}
\definecolor{diafillcolor}{rgb}{0.000000, 0.000000, 0.000000}
\pgfsetfillcolor{diafillcolor}
\pgfsetfillopacity{1.000000}
\node[anchor=base,inner sep=0pt, outer sep=0pt,color=dialinecolor] at (7.186250\du,35.291250\du){};
\definecolor{dialinecolor}{rgb}{0.000000, 0.000000, 0.000000}
\pgfsetstrokecolor{dialinecolor}
\pgfsetstrokeopacity{1.000000}
\definecolor{diafillcolor}{rgb}{0.000000, 0.000000, 0.000000}
\pgfsetfillcolor{diafillcolor}
\pgfsetfillopacity{1.000000}
\node[anchor=base,inner sep=0pt, outer sep=0pt,color=dialinecolor] at (7.186250\du,35.926250\du){Concerned list};
\pgfsetlinewidth{0.050000\du}
\pgfsetdash{}{0pt}
\pgfsetbuttcap
\pgfsetmiterjoin
\pgfsetlinewidth{0.050000\du}
\pgfsetbuttcap
\pgfsetmiterjoin
\pgfsetdash{}{0pt}
\definecolor{diafillcolor}{rgb}{1.000000, 1.000000, 1.000000}
\pgfsetfillcolor{diafillcolor}
\pgfsetfillopacity{1.000000}
\definecolor{dialinecolor}{rgb}{0.000000, 0.000000, 0.000000}
\pgfsetstrokecolor{dialinecolor}
\pgfsetstrokeopacity{1.000000}
\pgfpathmoveto{\pgfpoint{11.077100\du}{9.168084\du}}
\pgfpathcurveto{\pgfpoint{12.191687\du}{8.179521\du}}{\pgfpoint{12.748980\du}{7.850000\du}}{\pgfpoint{13.863567\du}{7.850000\du}}
\pgfpathcurveto{\pgfpoint{14.978153\du}{7.850000\du}}{\pgfpoint{15.535447\du}{8.179521\du}}{\pgfpoint{16.650033\du}{9.168084\du}}
\pgfpathlineto{\pgfpoint{16.650033\du}{14.440419\du}}
\pgfpathcurveto{\pgfpoint{15.535447\du}{15.428982\du}}{\pgfpoint{14.978153\du}{15.758503\du}}{\pgfpoint{13.863567\du}{15.758503\du}}
\pgfpathcurveto{\pgfpoint{12.748980\du}{15.758503\du}}{\pgfpoint{12.191687\du}{15.428982\du}}{\pgfpoint{11.077100\du}{14.440419\du}}
\pgfpathlineto{\pgfpoint{11.077100\du}{9.168084\du}}
\pgfpathclose
\pgfusepath{fill,stroke}
\pgfsetbuttcap
\pgfsetmiterjoin
\pgfsetdash{}{0pt}
\definecolor{dialinecolor}{rgb}{0.000000, 0.000000, 0.000000}
\pgfsetstrokecolor{dialinecolor}
\pgfsetstrokeopacity{1.000000}
\pgfpathmoveto{\pgfpoint{11.077100\du}{9.168084\du}}
\pgfpathcurveto{\pgfpoint{12.191687\du}{10.156647\du}}{\pgfpoint{12.748980\du}{10.486168\du}}{\pgfpoint{13.863567\du}{10.486168\du}}
\pgfpathcurveto{\pgfpoint{14.978153\du}{10.486168\du}}{\pgfpoint{15.535447\du}{10.156647\du}}{\pgfpoint{16.650033\du}{9.168084\du}}
\pgfusepath{stroke}
\definecolor{dialinecolor}{rgb}{0.000000, 0.000000, 0.000000}
\pgfsetstrokecolor{dialinecolor}
\pgfsetstrokeopacity{1.000000}
\definecolor{diafillcolor}{rgb}{0.000000, 0.000000, 0.000000}
\pgfsetfillcolor{diafillcolor}
\pgfsetfillopacity{1.000000}
\node[anchor=base,inner sep=0pt, outer sep=0pt,color=dialinecolor] at (13.863567\du,11.987043\du){Duplicates};
\definecolor{dialinecolor}{rgb}{0.000000, 0.000000, 0.000000}
\pgfsetstrokecolor{dialinecolor}
\pgfsetstrokeopacity{1.000000}
\definecolor{diafillcolor}{rgb}{0.000000, 0.000000, 0.000000}
\pgfsetfillcolor{diafillcolor}
\pgfsetfillopacity{1.000000}
\node[anchor=base,inner sep=0pt, outer sep=0pt,color=dialinecolor] at (13.863567\du,12.622043\du){};
\definecolor{dialinecolor}{rgb}{0.000000, 0.000000, 0.000000}
\pgfsetstrokecolor{dialinecolor}
\pgfsetstrokeopacity{1.000000}
\definecolor{diafillcolor}{rgb}{0.000000, 0.000000, 0.000000}
\pgfsetfillcolor{diafillcolor}
\pgfsetfillopacity{1.000000}
\node[anchor=base,inner sep=0pt, outer sep=0pt,color=dialinecolor] at (13.863567\du,13.257043\du){Removed};
\pgfsetlinewidth{0.050000\du}
\pgfsetdash{}{0pt}
\pgfsetbuttcap
\pgfsetmiterjoin
\pgfsetlinewidth{0.050000\du}
\pgfsetbuttcap
\pgfsetmiterjoin
\pgfsetdash{}{0pt}
\definecolor{diafillcolor}{rgb}{1.000000, 1.000000, 1.000000}
\pgfsetfillcolor{diafillcolor}
\pgfsetfillopacity{1.000000}
\definecolor{dialinecolor}{rgb}{0.000000, 0.000000, 0.000000}
\pgfsetstrokecolor{dialinecolor}
\pgfsetstrokeopacity{1.000000}
\pgfpathmoveto{\pgfpoint{23.622500\du}{9.283340\du}}
\pgfpathcurveto{\pgfpoint{24.742500\du}{8.585840\du}}{\pgfpoint{25.302500\du}{8.353340\du}}{\pgfpoint{26.422500\du}{8.353340\du}}
\pgfpathcurveto{\pgfpoint{27.542500\du}{8.353340\du}}{\pgfpoint{28.102500\du}{8.585840\du}}{\pgfpoint{29.222500\du}{9.283340\du}}
\pgfpathlineto{\pgfpoint{29.222500\du}{13.003340\du}}
\pgfpathcurveto{\pgfpoint{28.102500\du}{13.700840\du}}{\pgfpoint{27.542500\du}{13.933340\du}}{\pgfpoint{26.422500\du}{13.933340\du}}
\pgfpathcurveto{\pgfpoint{25.302500\du}{13.933340\du}}{\pgfpoint{24.742500\du}{13.700840\du}}{\pgfpoint{23.622500\du}{13.003340\du}}
\pgfpathlineto{\pgfpoint{23.622500\du}{9.283340\du}}
\pgfpathclose
\pgfusepath{fill,stroke}
\pgfsetbuttcap
\pgfsetmiterjoin
\pgfsetdash{}{0pt}
\definecolor{dialinecolor}{rgb}{0.000000, 0.000000, 0.000000}
\pgfsetstrokecolor{dialinecolor}
\pgfsetstrokeopacity{1.000000}
\pgfpathmoveto{\pgfpoint{23.622500\du}{9.283340\du}}
\pgfpathcurveto{\pgfpoint{24.742500\du}{9.980840\du}}{\pgfpoint{25.302500\du}{10.213340\du}}{\pgfpoint{26.422500\du}{10.213340\du}}
\pgfpathcurveto{\pgfpoint{27.542500\du}{10.213340\du}}{\pgfpoint{28.102500\du}{9.980840\du}}{\pgfpoint{29.222500\du}{9.283340\du}}
\pgfusepath{stroke}
\definecolor{dialinecolor}{rgb}{0.000000, 0.000000, 0.000000}
\pgfsetstrokecolor{dialinecolor}
\pgfsetstrokeopacity{1.000000}
\definecolor{diafillcolor}{rgb}{0.000000, 0.000000, 0.000000}
\pgfsetfillcolor{diafillcolor}
\pgfsetfillopacity{1.000000}
\node[anchor=base,inner sep=0pt, outer sep=0pt,color=dialinecolor] at (26.422500\du,10.814590\du){};
\definecolor{dialinecolor}{rgb}{0.000000, 0.000000, 0.000000}
\pgfsetstrokecolor{dialinecolor}
\pgfsetstrokeopacity{1.000000}
\definecolor{diafillcolor}{rgb}{0.000000, 0.000000, 0.000000}
\pgfsetfillcolor{diafillcolor}
\pgfsetfillopacity{1.000000}
\node[anchor=base,inner sep=0pt, outer sep=0pt,color=dialinecolor] at (26.422500\du,11.449590\du){Machine  translated};
\definecolor{dialinecolor}{rgb}{0.000000, 0.000000, 0.000000}
\pgfsetstrokecolor{dialinecolor}
\pgfsetstrokeopacity{1.000000}
\definecolor{diafillcolor}{rgb}{0.000000, 0.000000, 0.000000}
\pgfsetfillcolor{diafillcolor}
\pgfsetfillopacity{1.000000}
\node[anchor=base,inner sep=0pt, outer sep=0pt,color=dialinecolor] at (26.422500\du,12.084590\du){};
\definecolor{dialinecolor}{rgb}{0.000000, 0.000000, 0.000000}
\pgfsetstrokecolor{dialinecolor}
\pgfsetstrokeopacity{1.000000}
\definecolor{diafillcolor}{rgb}{0.000000, 0.000000, 0.000000}
\pgfsetfillcolor{diafillcolor}
\pgfsetfillopacity{1.000000}
\node[anchor=base,inner sep=0pt, outer sep=0pt,color=dialinecolor] at (26.422500\du,12.719590\du){lexicon};
\pgfsetlinewidth{0.050000\du}
\pgfsetdash{}{0pt}
\pgfsetbuttcap
\pgfsetmiterjoin
\pgfsetlinewidth{0.050000\du}
\pgfsetbuttcap
\pgfsetmiterjoin
\pgfsetdash{}{0pt}
\definecolor{diafillcolor}{rgb}{1.000000, 1.000000, 1.000000}
\pgfsetfillcolor{diafillcolor}
\pgfsetfillopacity{1.000000}
\definecolor{dialinecolor}{rgb}{0.000000, 0.000000, 0.000000}
\pgfsetstrokecolor{dialinecolor}
\pgfsetstrokeopacity{1.000000}
\pgfpathmoveto{\pgfpoint{12.600000\du}{16.850000\du}}
\pgfpathlineto{\pgfpoint{20.800000\du}{16.850000\du}}
\pgfpathlineto{\pgfpoint{19.160000\du}{20.100000\du}}
\pgfpathlineto{\pgfpoint{14.240000\du}{20.100000\du}}
\pgfpathlineto{\pgfpoint{12.600000\du}{16.850000\du}}
\pgfpathclose
\pgfusepath{fill,stroke}
\definecolor{dialinecolor}{rgb}{0.000000, 0.000000, 0.000000}
\pgfsetstrokecolor{dialinecolor}
\pgfsetstrokeopacity{1.000000}
\definecolor{diafillcolor}{rgb}{0.000000, 0.000000, 0.000000}
\pgfsetfillcolor{diafillcolor}
\pgfsetfillopacity{1.000000}
\node[anchor=base,inner sep=0pt, outer sep=0pt,color=dialinecolor] at (16.700000\du,17.998750\du){Manually evaluate };
\definecolor{dialinecolor}{rgb}{0.000000, 0.000000, 0.000000}
\pgfsetstrokecolor{dialinecolor}
\pgfsetstrokeopacity{1.000000}
\definecolor{diafillcolor}{rgb}{0.000000, 0.000000, 0.000000}
\pgfsetfillcolor{diafillcolor}
\pgfsetfillopacity{1.000000}
\node[anchor=base,inner sep=0pt, outer sep=0pt,color=dialinecolor] at (16.700000\du,18.633750\du){};
\definecolor{dialinecolor}{rgb}{0.000000, 0.000000, 0.000000}
\pgfsetstrokecolor{dialinecolor}
\pgfsetstrokeopacity{1.000000}
\definecolor{diafillcolor}{rgb}{0.000000, 0.000000, 0.000000}
\pgfsetfillcolor{diafillcolor}
\pgfsetfillopacity{1.000000}
\node[anchor=base,inner sep=0pt, outer sep=0pt,color=dialinecolor] at (16.700000\du,19.268750\du){the translated list};
\pgfsetlinewidth{0.050000\du}
\pgfsetdash{}{0pt}
\pgfsetbuttcap
{
\definecolor{diafillcolor}{rgb}{0.000000, 0.000000, 0.000000}
\pgfsetfillcolor{diafillcolor}
\pgfsetfillopacity{1.000000}
\pgfsetarrowsend{to}
\definecolor{dialinecolor}{rgb}{0.000000, 0.000000, 0.000000}
\pgfsetstrokecolor{dialinecolor}
\pgfsetstrokeopacity{1.000000}
\draw (7.587332\du,8.081583\du)--(7.769505\du,6.824144\du);
}
\pgfsetlinewidth{0.050000\du}
\pgfsetdash{}{0pt}
\pgfsetbuttcap
{
\definecolor{diafillcolor}{rgb}{0.000000, 0.000000, 0.000000}
\pgfsetfillcolor{diafillcolor}
\pgfsetfillopacity{1.000000}
\pgfsetarrowsend{to}
\definecolor{dialinecolor}{rgb}{0.000000, 0.000000, 0.000000}
\pgfsetstrokecolor{dialinecolor}
\pgfsetstrokeopacity{1.000000}
\draw (9.773699\du,6.824425\du)--(11.424710\du,8.834698\du);
}
\pgfsetlinewidth{0.050000\du}
\pgfsetdash{}{0pt}
\pgfsetbuttcap
{
\definecolor{diafillcolor}{rgb}{0.000000, 0.000000, 0.000000}
\pgfsetfillcolor{diafillcolor}
\pgfsetfillopacity{1.000000}
\pgfsetarrowsend{to}
\definecolor{dialinecolor}{rgb}{0.000000, 0.000000, 0.000000}
\pgfsetstrokecolor{dialinecolor}
\pgfsetstrokeopacity{1.000000}
\draw (15.576257\du,8.292321\du)--(16.194142\du,7.025326\du);
\draw (21.576257\du,8.67\du)--(23.3\du,7.08\du);
;
}
\pgfsetlinewidth{0.050000\du}
\pgfsetdash{}{0pt}
\pgfsetbuttcap
{
\definecolor{diafillcolor}{rgb}{0.000000, 0.000000, 0.000000}
\pgfsetfillcolor{diafillcolor}
\pgfsetfillopacity{1.000000}
\pgfsetarrowsend{to}
\definecolor{dialinecolor}{rgb}{0.000000, 0.000000, 0.000000}
\pgfsetstrokecolor{dialinecolor}
\pgfsetstrokeopacity{1.000000}
\draw (18.268036\du,7.024347\du)--(19.009251\du,8.615265\du);
}
\pgfsetlinewidth{0.050000\du}
\pgfsetdash{}{0pt}
\pgfsetbuttcap
{
\definecolor{diafillcolor}{rgb}{0.000000, 0.000000, 0.000000}
\pgfsetfillcolor{diafillcolor}
\pgfsetfillopacity{1.000000}
\pgfsetarrowsend{to}
\definecolor{dialinecolor}{rgb}{0.000000, 0.000000, 0.000000}
\pgfsetstrokecolor{dialinecolor}
\pgfsetstrokeopacity{1.000000}
\draw (26.215575\du,7.074347\du)--(26.279524\du,8.331855\du);
}
\pgfsetlinewidth{0.050000\du}
\pgfsetdash{}{0pt}
\pgfsetbuttcap
{
\definecolor{diafillcolor}{rgb}{0.000000, 0.000000, 0.000000}
\pgfsetfillcolor{diafillcolor}
\pgfsetfillopacity{1.000000}
\pgfsetarrowsend{to}
\definecolor{dialinecolor}{rgb}{0.000000, 0.000000, 0.000000}
\pgfsetstrokecolor{dialinecolor}
\pgfsetstrokeopacity{1.000000}
\draw (23.785070\du,13.132205\du)--(18.886138\du,16.826450\du);
}
\pgfsetlinewidth{0.050000\du}
\pgfsetdash{}{0pt}
\pgfsetbuttcap
{
\definecolor{diafillcolor}{rgb}{0.000000, 0.000000, 0.000000}
\pgfsetfillcolor{diafillcolor}
\pgfsetfillopacity{1.000000}
\pgfsetarrowsend{to}
\definecolor{dialinecolor}{rgb}{0.000000, 0.000000, 0.000000}
\pgfsetstrokecolor{dialinecolor}
\pgfsetstrokeopacity{1.000000}
\draw (16.682147\du,20.124704\du)--(16.650000\du,23.095300\du);
}
\pgfsetlinewidth{0.050000\du}
\pgfsetdash{}{0pt}
\pgfsetbuttcap
{
\definecolor{diafillcolor}{rgb}{0.000000, 0.000000, 0.000000}
\pgfsetfillcolor{diafillcolor}
\pgfsetfillopacity{1.000000}
\pgfsetarrowsend{to}
\definecolor{dialinecolor}{rgb}{0.000000, 0.000000, 0.000000}
\pgfsetstrokecolor{dialinecolor}
\pgfsetstrokeopacity{1.000000}
\draw (20.559400\du,25.050000\du)--(24.409515\du,24.920609\du);
}
\pgfsetlinewidth{0.050000\du}
\pgfsetdash{}{0pt}
\pgfsetbuttcap
{
\definecolor{diafillcolor}{rgb}{0.000000, 0.000000, 0.000000}
\pgfsetfillcolor{diafillcolor}
\pgfsetfillopacity{1.000000}
\pgfsetarrowsend{to}
\definecolor{dialinecolor}{rgb}{0.000000, 0.000000, 0.000000}
\pgfsetstrokecolor{dialinecolor}
\pgfsetstrokeopacity{1.000000}
\draw (12.740600\du,25.050000\du)--(9.647474\du,25.116612\du);
}
\pgfsetlinewidth{0.050000\du}
\pgfsetdash{}{0pt}
\pgfsetbuttcap
{
\definecolor{diafillcolor}{rgb}{0.000000, 0.000000, 0.000000}
\pgfsetfillcolor{diafillcolor}
\pgfsetfillopacity{1.000000}
\pgfsetarrowsend{to}
\definecolor{dialinecolor}{rgb}{0.000000, 0.000000, 0.000000}
\pgfsetstrokecolor{dialinecolor}
\pgfsetstrokeopacity{1.000000}
\draw (16.660339\du,27.023654\du)--(16.666102\du,28.124724\du);
}
\pgfsetlinewidth{0.050000\du}
\pgfsetdash{}{0pt}
\pgfsetbuttcap
{
\definecolor{diafillcolor}{rgb}{0.000000, 0.000000, 0.000000}
\pgfsetfillcolor{diafillcolor}
\pgfsetfillopacity{1.000000}
\pgfsetarrowsend{to}
\definecolor{dialinecolor}{rgb}{0.000000, 0.000000, 0.000000}
\pgfsetstrokecolor{dialinecolor}
\pgfsetstrokeopacity{1.000000}
\draw (14.240000\du,20.100000\du)--(10.147166\du,19.984790\du);
}
\pgfsetlinewidth{0.050000\du}
\pgfsetdash{}{0pt}
\pgfsetbuttcap
{
\definecolor{diafillcolor}{rgb}{0.000000, 0.000000, 0.000000}
\pgfsetfillcolor{diafillcolor}
\pgfsetfillopacity{1.000000}
\pgfsetarrowsend{to}
\definecolor{dialinecolor}{rgb}{0.000000, 0.000000, 0.000000}
\pgfsetstrokecolor{dialinecolor}
\pgfsetstrokeopacity{1.000000}
\draw (13.016516\du,34.472888\du)--(9.997716\du,34.409259\du);
}
\pgfsetlinewidth{0.050000\du}
\pgfsetdash{}{0pt}
\pgfsetbuttcap
{
\definecolor{diafillcolor}{rgb}{0.000000, 0.000000, 0.000000}
\pgfsetfillcolor{diafillcolor}
\pgfsetfillopacity{1.000000}
\pgfsetarrowsend{to}
\definecolor{dialinecolor}{rgb}{0.000000, 0.000000, 0.000000}
\pgfsetstrokecolor{dialinecolor}
\pgfsetstrokeopacity{1.000000}
\draw (20.334187\du,34.468564\du)--(24.077312\du,34.385260\du);
}
\pgfsetlinewidth{0.050000\du}
\pgfsetdash{}{0pt}
\pgfsetbuttcap
{
\definecolor{diafillcolor}{rgb}{0.000000, 0.000000, 0.000000}
\pgfsetfillcolor{diafillcolor}
\pgfsetfillopacity{1.000000}
\pgfsetarrowsend{to}
\definecolor{dialinecolor}{rgb}{0.000000, 0.000000, 0.000000}
\pgfsetstrokecolor{dialinecolor}
\pgfsetstrokeopacity{1.000000}
\draw (16.674994\du,31.500264\du)--(16.674990\du,32.724773\du);
}
\definecolor{dialinecolor}{rgb}{0.000000, 0.000000, 0.000000}
\pgfsetstrokecolor{dialinecolor}
\pgfsetstrokeopacity{1.000000}
\definecolor{diafillcolor}{rgb}{0.000000, 0.000000, 0.000000}
\pgfsetfillcolor{diafillcolor}
\pgfsetfillopacity{1.000000}
\node[anchor=base west,inner sep=0pt,outer sep=0pt,color=dialinecolor] at (21.205000\du,24.575000\du){Not-correct};
\definecolor{dialinecolor}{rgb}{0.000000, 0.000000, 0.000000}
\pgfsetstrokecolor{dialinecolor}
\pgfsetstrokeopacity{1.000000}
\definecolor{diafillcolor}{rgb}{0.000000, 0.000000, 0.000000}
\pgfsetfillcolor{diafillcolor}
\pgfsetfillopacity{1.000000}
\node[anchor=base west,inner sep=0pt,outer sep=0pt,color=dialinecolor] at (10.405000\du,24.625000\du){Undecided};
\definecolor{dialinecolor}{rgb}{0.000000, 0.000000, 0.000000}
\pgfsetstrokecolor{dialinecolor}
\pgfsetstrokeopacity{1.000000}
\definecolor{diafillcolor}{rgb}{0.000000, 0.000000, 0.000000}
\pgfsetfillcolor{diafillcolor}
\pgfsetfillopacity{1.000000}
\node[anchor=base west,inner sep=0pt,outer sep=0pt,color=dialinecolor] at (17.355000\du,27.525000\du){Correct};
\end{tikzpicture}